\documentclass[times,twocolumn,final]{article}
\usepackage{spconf,amsmath,graphicx}
\usepackage{fixltx2e}
\usepackage{mathtools}
\usepackage{threeparttable}
\usepackage{multirow}
\usepackage{tabularx,ragged2e}
\usepackage{array}
\usepackage{framed,multirow}
\usepackage{amssymb}
\usepackage{latexsym}
\usepackage{url}
\usepackage{xcolor}
\usepackage{algorithm}
\usepackage[noend]{algpseudocode}
\usepackage{siunitx}

\makeatletter
\def\BState{\State\hskip-\ALG@thistlm}
\makeatother

\def\x{{\mathbf x}}

\DeclarePairedDelimiter\abs{\lvert}{\rvert}%
\DeclarePairedDelimiter\norm{\lVert}{\rVert}%

\makeatletter
\let\oldabs\abs
\def\abs{\@ifstar{\oldabs}{\oldabs*}}

\let\oldnorm\norm
\def\norm{\@ifstar{\oldnorm}{\oldnorm*}}
\makeatother

\title{Deep Deformable Registration: Enhancing Accuracy by Fully Convolutional Neural Net}

\name{Sayan Ghosal $^{\star}$ \quad Nilanjan Ray $^{\dagger}$}

\address{$^{\star}$ Department of Electronics and Telecom. Engineering, Jadavpur University, Kolkata, India. \\ 
     $^{\dagger}$ Department of Computing Science, University of Alberta, Canada.\\
     (Corresponding author's e-mail: nray1@ualberta.ca)}
										
\begin{document}
%
\maketitle
\begin{abstract}
Deformable registration is ubiquitous in medical image analysis. Many deformable registration methods minimize sum of squared difference (SSD) as the registration cost with respect to deformable model parameters. In this work, we construct a tight upper bound of the SSD registration cost by using a fully convolutional neural network (FCNN) in the registration pipeline. The upper bound SSD (UB-SSD) enhances the original deformable model parameter space by adding a heatmap output from FCNN. Next, we minimize this UB-SSD by adjusting both the parameters of the FCNN and the parameters of the deformable model in coordinate descent. Our coordinate descent framework is end-to-end and can work with any deformable registration method that uses SSD. We demonstrate experimentally that our method enhances the accuracy of deformable registration algorithms significantly on two publicly available 3D brain MRI data sets.
\end{abstract}




\section{Introduction}
\label{sec:intro}

Image registration or image alignment is the process of overlaying two images taken at different time instants, or different view points, or from different subjects in a common coordinate system. Image registration has remained a significant tool in medical imaging applications \cite{R13}. 3D data in medical images, where image registration is applied, generally includes Computed Tomography (CT), Cone-beam CT (CBCT), Magnetic Resonance Imaging (MRI) and Computer Aided Design (CAD) model of medical devices.

Among all the image registration methods, deformable image registration is important in neuroscience and clinical studies. Diffeomorphic demons \cite{TV2009} and Log-domain diffeomorphic demon \cite{TV2008} algorithms are popular deformable registration methods. In these optimization-based methods, the deformable transformation parameters are iteratively optimized over a scalar valued cost (e.g., SSD) representing the quality of registration \cite{ACN2013}. To impose smoothness on the solution, typically a regularization term is also added to the registration cost function. These costs are non-convex in nature, hence the optimization sometimes get trapped in the local minima. On the basis of the cost functions, different optimization algorithms are used. In the Gauss-Newton method for minimizing SSD, projective geometric deformation is used \cite{RM97}. The method is sensitive to local minima. Levenberg-Marquadt algorithm was used in \cite{HR99} to minimize the difference in intensities of corresponding pixels. This method update the parameters between gradient descent and Gauss Newton and accelerates towards local minima. The combination of Levenber-Marquadt method and SSD is used in \cite{PUM95}.

In order to improve the solution of registration, in other words, to find better local minima in the SSD cost, we propose a novel method to modify the reference image by introducing a heatmap (essentially another image) produced by a Fully Convolutional Neural Network (FCNN) with a skip architecture \cite{JET15}. This modified reference image helps to create a tight upper bound to the SSD registration cost that we refer to as UB-SSD. Next, we minimize the UB-SSD by adjusting parameters of the FCNN as well as deformation parameters of the registration algorithm. We refer to our proposed method by deep deformable registration (DDR).

FCNN is type of deep learning machine that has has been successfully used for semantic segmentation in \cite{JET15}. In \cite{KAG2012} convolutional network has been used to classify 1.2 million images in 1000 different classes. However, our proposed method (DDR) does not employ any learning, rather it uses FCNN to optimize SSD registration cost. It is a newer trend in computer vision and graphics, where deep learning tools are used only for optimization purposes and not for learning. One prominent example is artistic style transfer \cite{LASM15}.

Prior to our work, convolutional network was used for image registration in \cite{Wu2013}, where the authors trained the parameters of a 2-layers convolutional network. The network was used to seek the hierarchical representation for each image, where high level features are inferred from the low level network. The goal of their work was to learn feature vectors for better registration. In contrast, DDR focuses on finding better solution during optimization, where we make use of end-to-end back-propagation and the deep learning architecture.


\section{Proposed Method}
In this section, we provide detailed descriptions of each component of our solution pipeline. We start with the overall registration framework. Then, we illustrate our FCNN architecture and define upper bound of the SSD cost. We end the section by explaining the registration module.

\subsection{Deep Deformable Registration Framework}
\begin{figure}[h]
  \centering
     \includegraphics[width=\columnwidth, height = 1.6in]{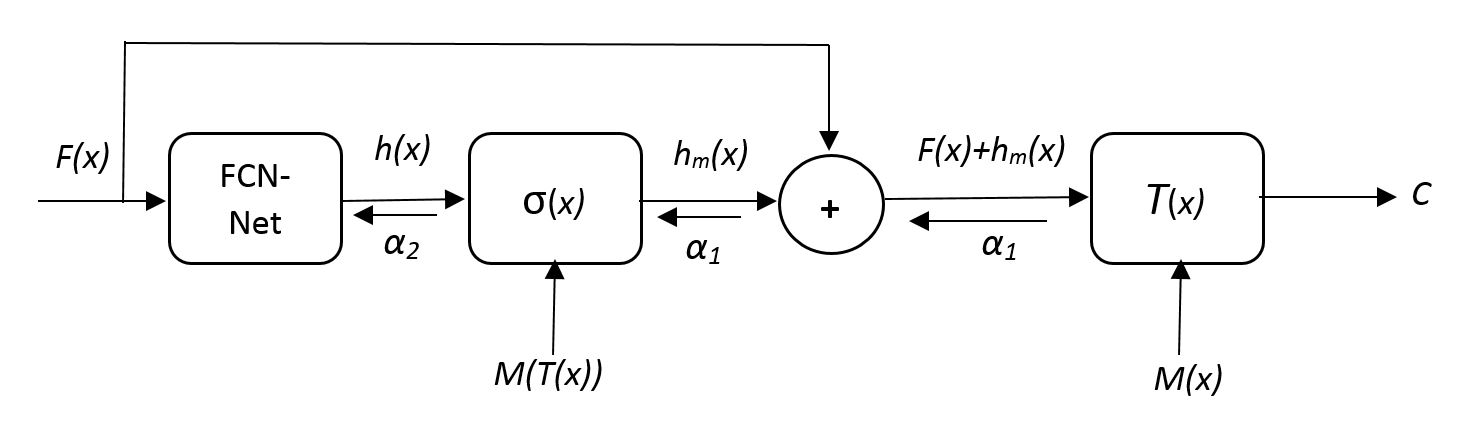}
\caption{Deep deformable registration framework.}
\label{fig1}
\end{figure}

Throughout this paper the moving image is represented as $M(x)$ and the fixed (or reference) image is represented as $F(x)$. The output of the fully convolutional neural network is represented as $h(x)$. $T(x)$ represents the deformation produced by a registration algorithm. The complete DDR framework is shown in Fig. \ref{fig1}.

In DDR, we have considered SSD as the registration cost between the moving and the fixed image. For simplicity, we omit any regularization term here. Hence, the SSD registration cost is as follows:

\begin{equation}
C = \frac {1}{2} \int_{}^{}[F(x)-M(T(x))]^2dx.
\label{eq1}
\end{equation}

In DDR, the FCNN is followed by a non-linear function $\sigma$ as shown in Fig. 1. A suitable design of $\sigma$ ensures the upper bound of the original cost (UB-SSD):

\begin{equation}
C_U = \frac {1}{2} \int_{}^{}[F(x)+h_m(x)-M(T(x))]^2dx.
\label{eq2}
\end{equation}

In order to minimize the UB-SSD given in (\ref{eq2}), back-propagation is applied. $\alpha_1$ is error that back-propagates from registration module and $\alpha_2$ is the error of the non-linear module. The output of the FCNN is the heatmap $h(x),$ which is modified by the non-linearity function $\sigma$ to $h_m(x)$:
\begin{equation}
h_m(x) = \sigma(h(x)).
\label{eq1a}
\end{equation}

\noindent This modified heat map is added pixel-wise with the fixed image $F(x)$ as a distortion. 

In DDR, the registration module minimizes the UB-SSD, which will ensure minimization of the original SSD cost. The DDR framework works in an iterative coordinate descent manner by alternating between the following two steps until convergence: (a) fix $h(x)$ and optimize for deformable parameters $T(x)$ and (b) fix $T(x)$ and optimize for heatmap $h(x)$ by back-propagation. Thus, the DDR framework works in an end-to-end fashion. The error signals $\alpha_1$ and $\alpha_2$ are as follows:
\begin{equation}
\alpha_2 = \sigma'(h(x))\alpha_1,
\label{eq3}
\end{equation}
\noindent and
\begin{equation}
\alpha_1 = \nabla_{h_m} {C_U} = F(x)+h_m(x)-M(T(x)).
\label{eq4}
\end{equation}

\begin{figure}[H]
  \centering
     \includegraphics[width=\columnwidth]{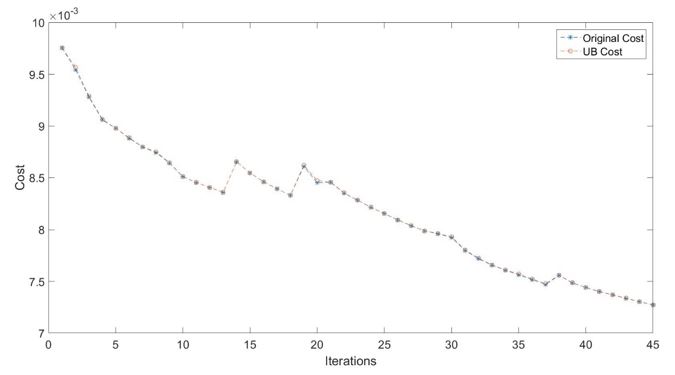}
\caption{ Original cost and UB-cost vs iterations.}
\label{fig2}
\end{figure}

Thus, the DDR framework enhances the space of optimization parameters from $T$ to a joint space of $T$ and $h$ (or $h_m$). So, when the registration optimizer is stuck at a local minimum of $T$, the alternating coordinate descent finds a better minimum for $h$ in the joint space, and the registration proceeds because of the upper bound property of the cost function. The decrease of UB-SSD and SSD is shown in Fig. \ref{fig2} over iterations of the coordinate descent for a registration example.

\subsection{Tight UB-SSD}
In DDR, (\ref{eq2}) serves as an upper bound to (\ref{eq1}). Using this condition, we obtain: 
\begin{equation}
\int_{}^{}\sigma(h(x))[\sigma(h(x))+2(F(x)-M(T(x)))]dx\geq 0.
\label{eq5}
\end{equation}

\noindent We ensure condition (\ref{eq5}) by realizing $\sigma$ as a soft thresholding function with threshold $t$:
\begin{equation}
  \sigma(z)=\left\{
  \begin{array}{@{}ll@{}}
    z-t, & \text{if}\ t\ge z \\
    0, & \text{if}\ -t\leq z \leq +t \\
		z+t, & \text{if}\ z\le -t. 
  \end{array}\right.
	\label{eq6}
\end{equation} 

Note that $\sigma$ is applied pixel-wise on the heatmap $h(x)$. For a tight UB-SSD condition (\ref{eq5}) can be restated as follows:
\begin{equation}
\epsilon \geq \int_{}^{}[\sigma(h(x))[\sigma(h(x))+2(F(x)-M(T(x)))]dx \geq 0,
\label{eq7}
\end{equation}

\noindent where $\epsilon$ is a small positive number. The following simple algorithm makes sure that with a soft thresholding function $\sigma$, condition (\ref{eq7}) is met. Fig. \ref{fig2} demonstrates UB-SSD is quite tight on the SSD cost.
\begin{algorithm}
\caption{Soft thresholding algorithm}\label{sth}
\begin{algorithmic}[1]
\State $\textit{t} \gets 0$
\State $ \textit{Set stepsize to a very small number}$
\BState \emph{loop}:
\If {$ \textit{Condition $(\ref{eq7})$ is not met}$}
\EndIf
\State $t \gets t+\textit{stepsize}$
\State \textbf{goto} \emph{loop}.
\end{algorithmic}
\end{algorithm}

\subsection{FCNN Architecture}
\begin{figure}[!htb]
  \centering
     \includegraphics[scale=0.42]{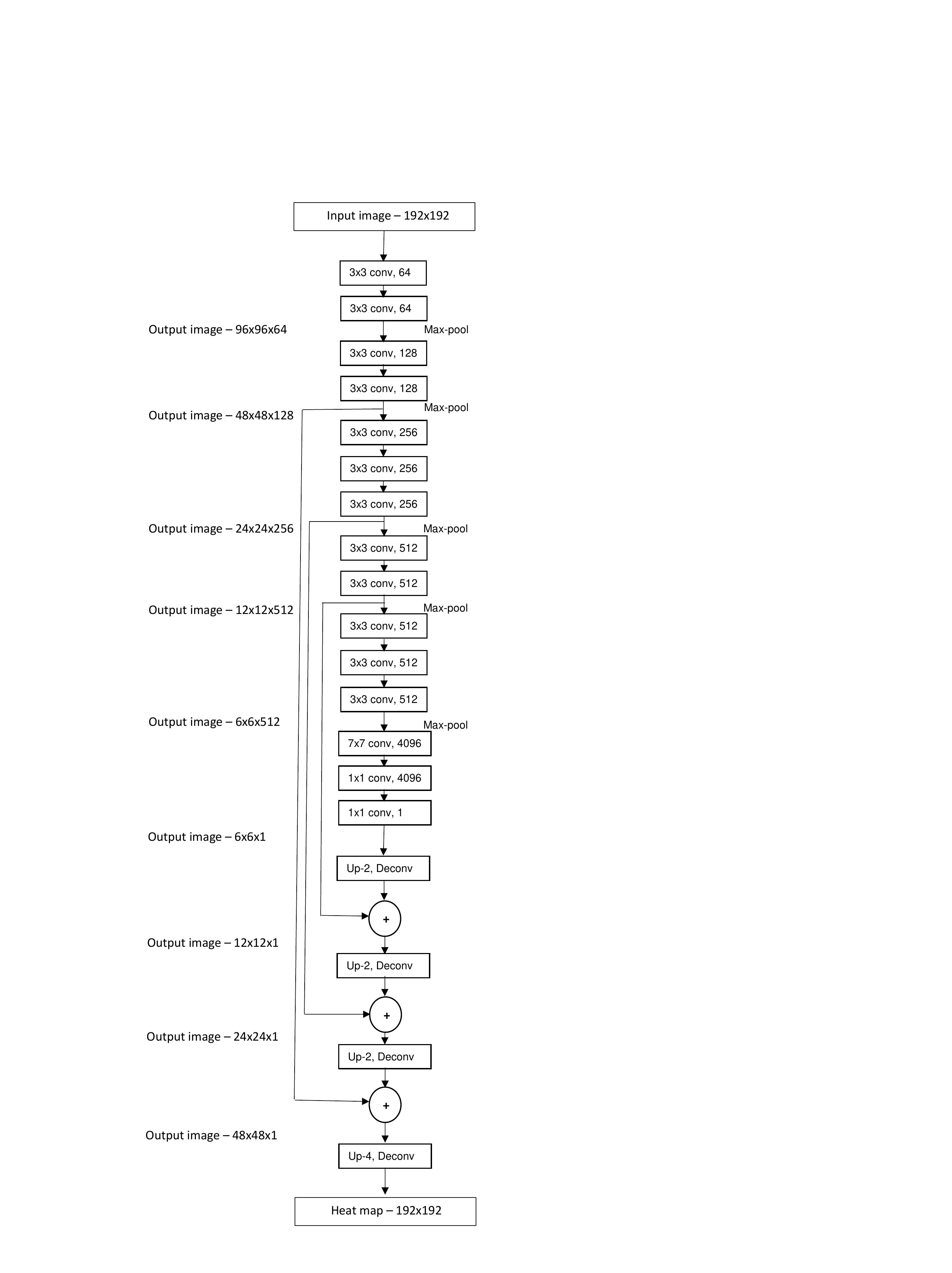}
\caption{FCNN with skip architecture.}
\label{fig3}
\end{figure}

We have used VGG-net \cite{KA15} to serve as FCNN. This network contains both convolutional and deconvolutional layers. In the VGG-net we decapitated the final classifier layer and convert all fully connected layers to convolutional layers. This is followed by multiple deconvolutional layers to bi-linearly up-sample the coarse outputs to pixel dense outputs. The convolutional part consist of multiple convolutional layers, ReLU layers and max-pooling layers. The deconvolutional part consists of deconvolutional layers. We have skipped multiple layers and fused them with the deconvolved layers to introduce local appearance information. A typical skip architecture used in our module is shown in details in Fig. \ref{fig3}.

\subsection{Registration module}
In order to register the moving image with the modified reference image we have used the demons \cite{TV2009, TV2008} method. To find the optimum transformation $T,$ we optimize the following cost:
\begin{equation}
E(T) = \frac{1}{2} \int_{}^{}[f(x)+ h_m(x)-M(T(x))]^2+\|\nabla T(x)\|^2]dx.
\label{eq8}
\end{equation}
Due to the large number of transformation parameters in the transformation field $T$, we use limited memory BFGS (LBFGS) algorithm to find the optimum transformation field. This algorithm is computationally less extensive than BFGS when the number of optimization parameters are large. While calculating the step length and direction, LBFGS store the hessian matrix for the last few iterations and use them to calculate the direction and step length instead of updating and storing the hessian matrix in each iteration. After finding the optimum transformation field, the error is back-propagated through the pipeline which helps the FCNN in finding the necessary distortion required to reduce the energy further down.


\section{Results}
\subsection{Registration Algorithms}
To establish the usefulness of DDR, the following two deformable registration algorithms, each with and without DDR, are used: 
\begin{enumerate}
 \item DDR + Diffeomorphic demon
 \item Diffeomorphic demon
 \item DDR + Log-demon
 \item Log-demon.
\end{enumerate}
In our setup, to register images using DDR + diffeomorphic demons, we have used FCNN-16s network \cite{JET15} and for registration using DDR + log-demon, we have used FCNN-32s architecture for the FCNN.

\subsection{Registration Evaluation Metrics}
For performance measures, we have used structural similarity index (SSIM) \cite{ZAH2004}, Peak signal to noise ratio (PSNR) and the SSD error. SSIM can capture local differences between the images, whereas SSD and PSNR can capture global differences.

\subsection{Experiments with IXI Dataset}
IXI dataset (\url{http://biomedic.doc.ic.ac.uk/brain-development/index.php?n=Main.Datasets}) consists of 30 subjects, which are all 3D volumetric data. Among them, we have randomly chosen one as our reference volume and we have registered the others using the aforementioned four algorithms.
\begin{figure}[!htb]
  \centering
	  \hspace{0.1 in}
		
     \includegraphics[width=3in, height = 2.4in]{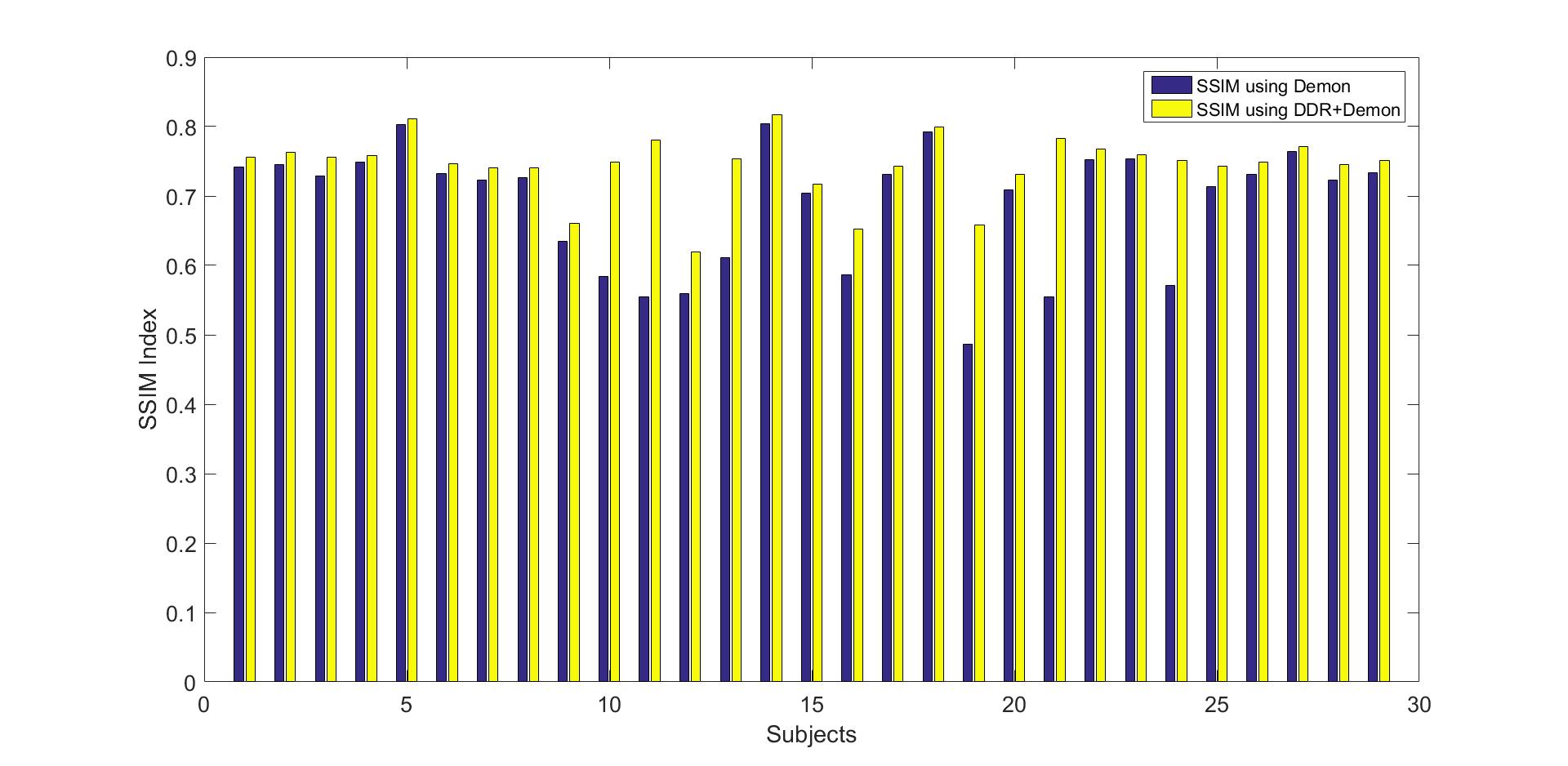}
	
		 \vspace{0.05 in}
		
     \includegraphics[width=3in, height =2.4in]{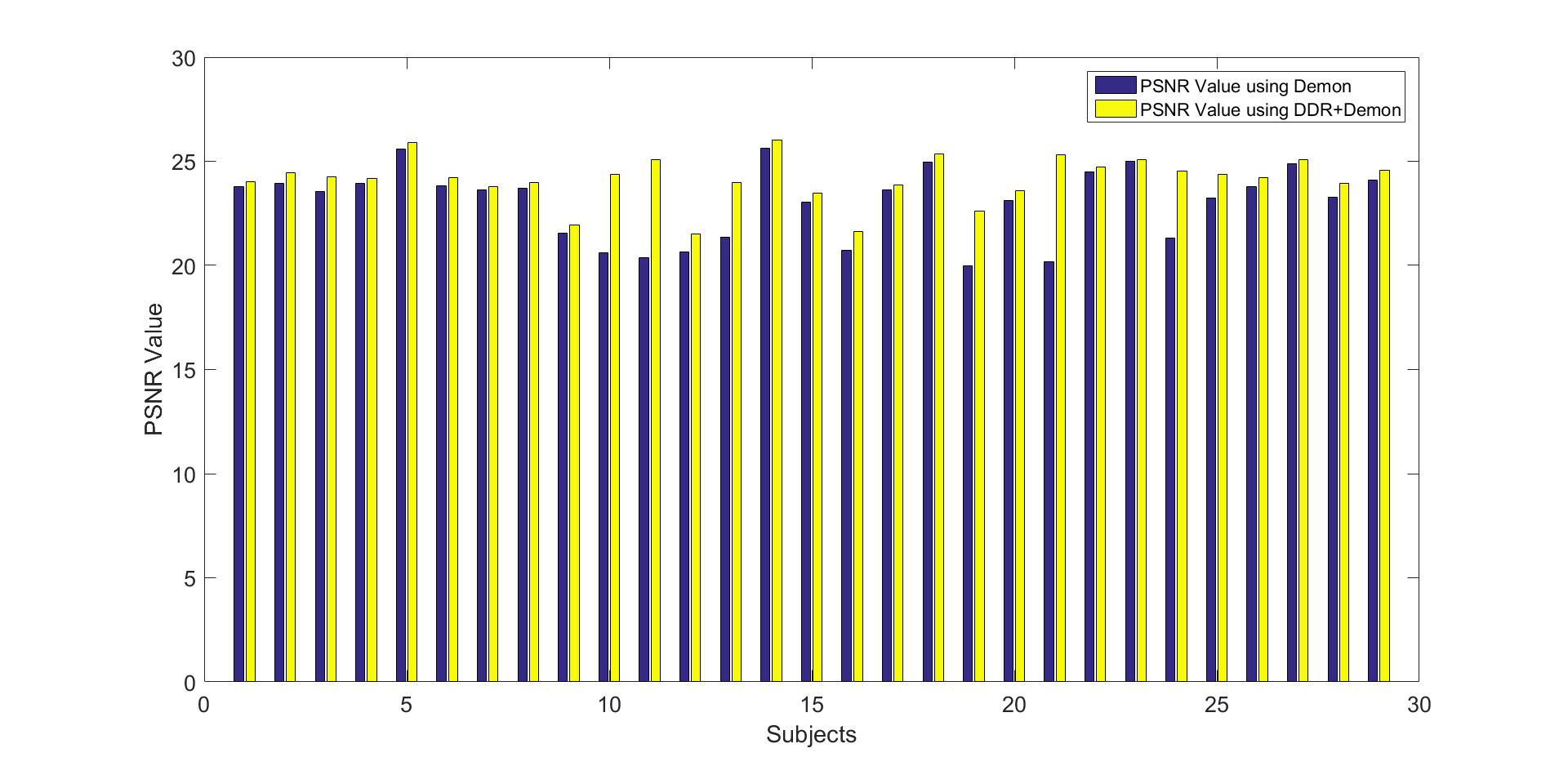}
		 \hspace{0.1 in}
     \includegraphics[width=3in, height =2.4in]{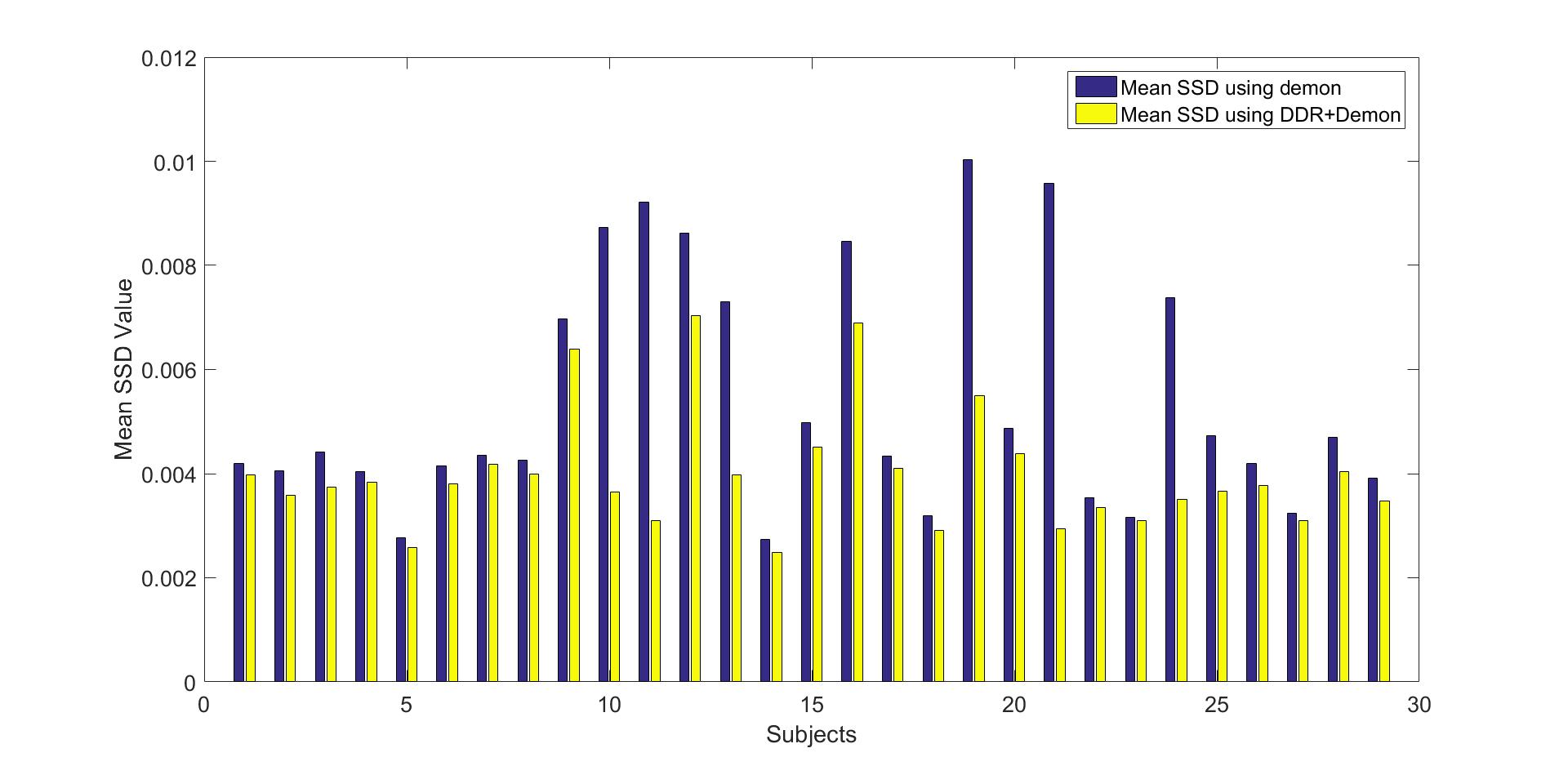} 
     \vspace{0.02in}
\caption{Results on IXI Dataset with diffeomorphic demons. Top : SSIM vs no. of Subjects, Middle: PSNR value vs no. of subjects, Bottom: Mean SSD value vs no. of subjects.}
\label{fig4}
\end{figure}
 
\begin{figure}[!htb]
  \centering
	  \hspace{0.6 in}
		
     \includegraphics[width=3.2in, height = 2.4in]{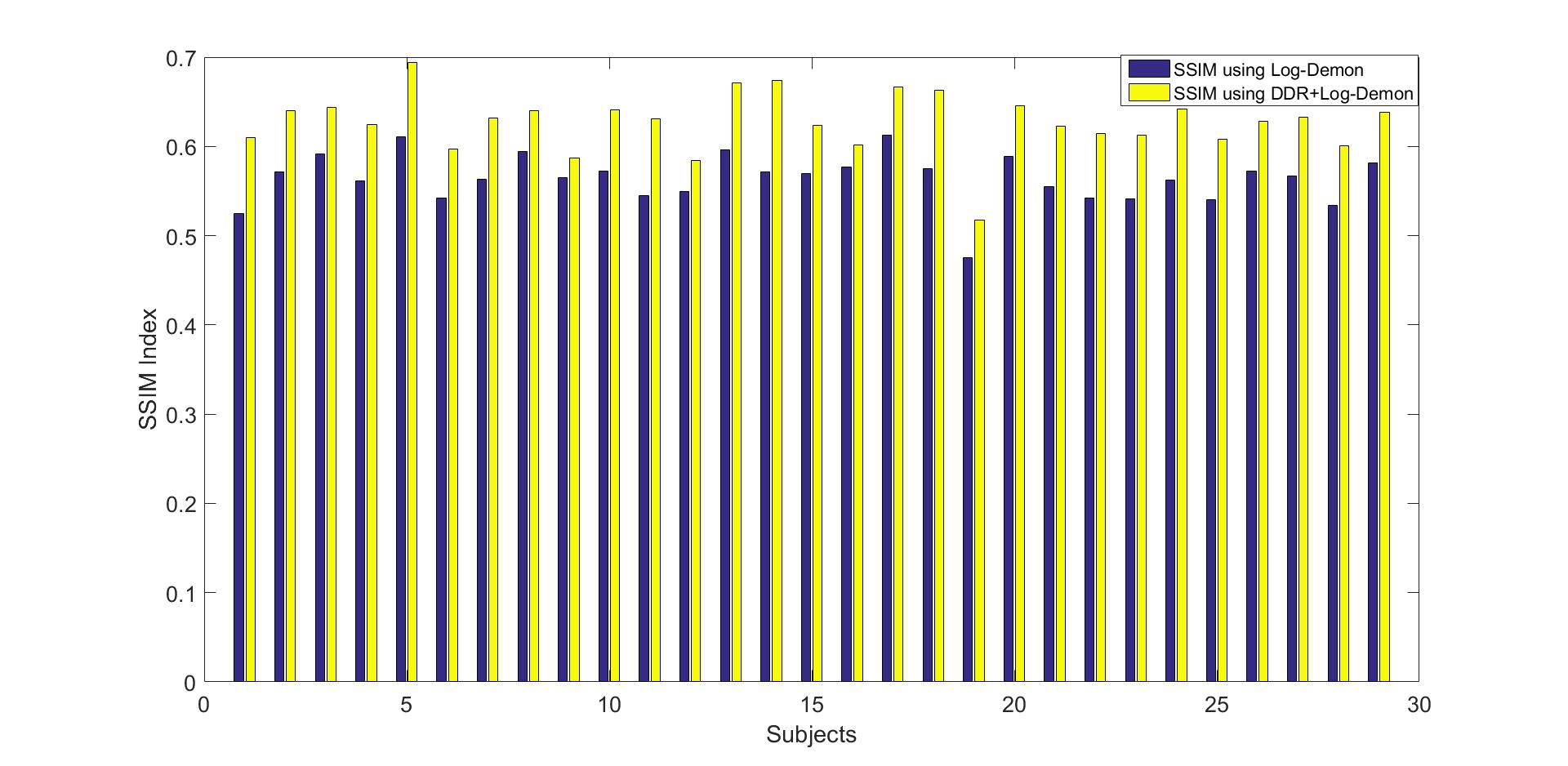}
	
		 \vspace{0.05 in}
		
     \includegraphics[width=3.2in, height =2.4in]{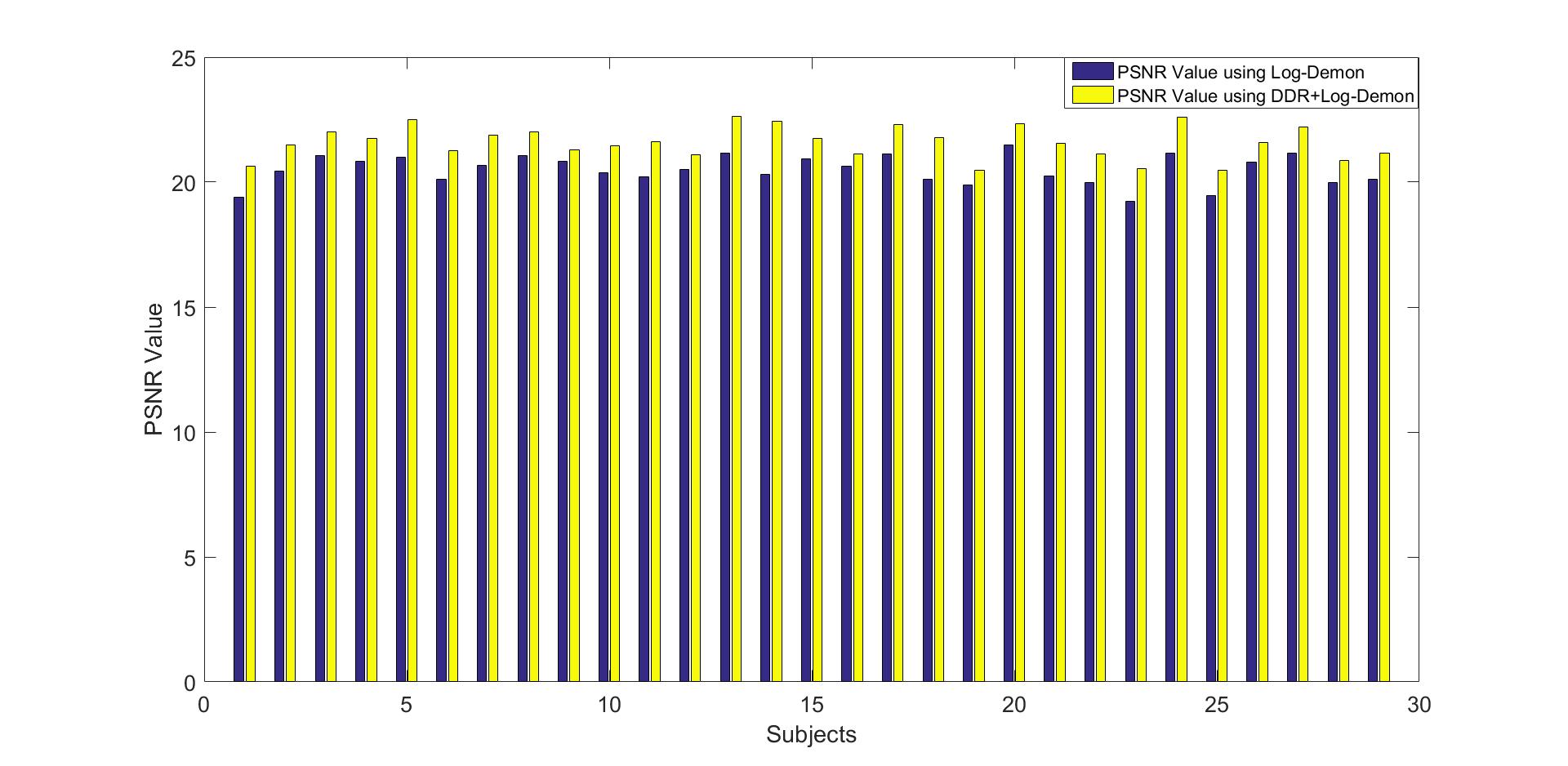}
		 \hspace{0.1 in}
     \includegraphics[width=3.2in, height =2.4in]{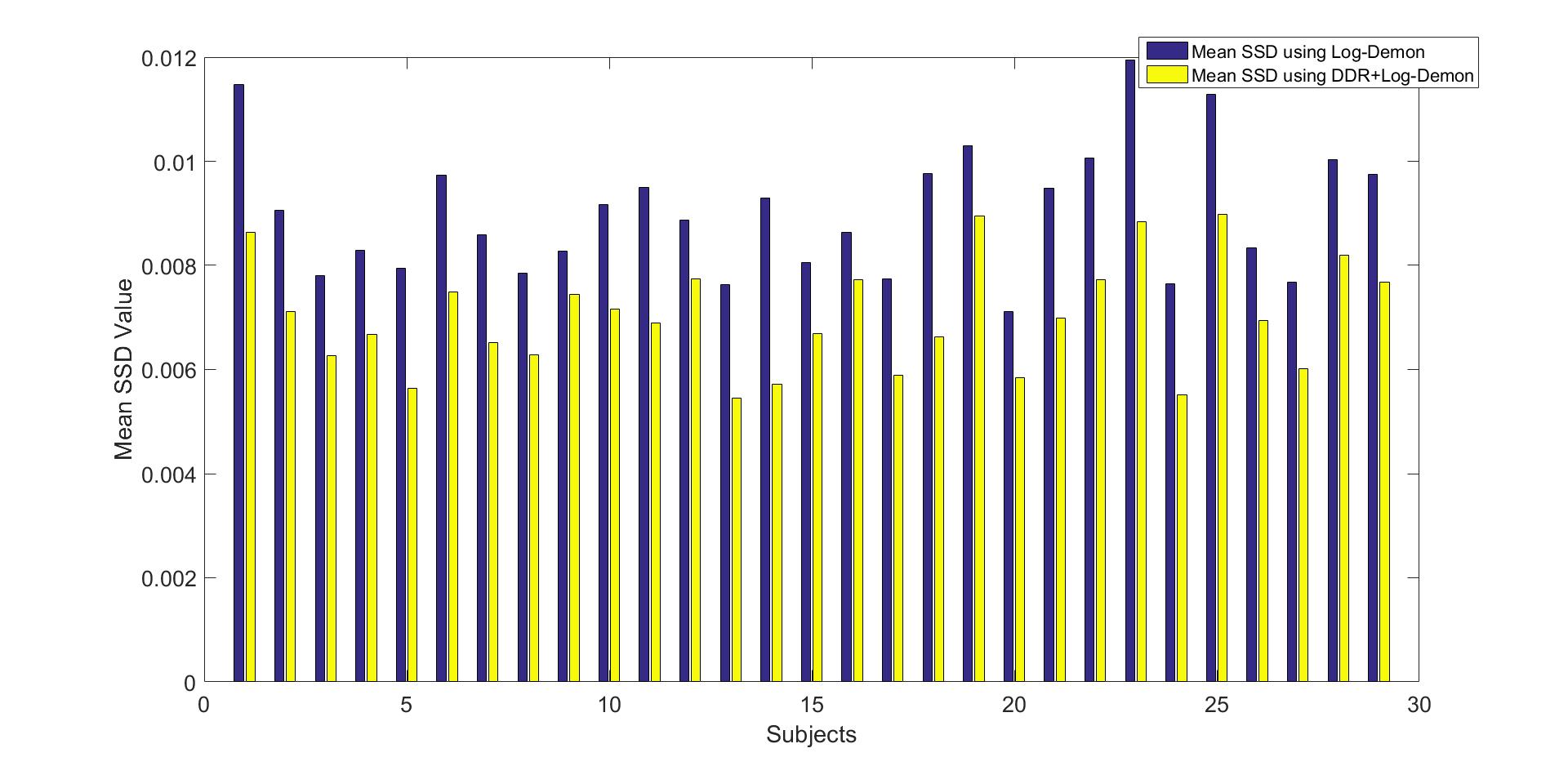} 
     \vspace{0.02in}
\caption{Results on IXI Dataset using log-demon. Top : SSIM vs no. of Subjects, Middle: PSNR value vs no. of subjects, Bottom: Mean SSD value vs no. of subjects.}
\label{fig5}
\end{figure}

The improvements in SSIM, PSNR and SSD error with diffeomorphic demon are provided in Fig. \ref{fig4}. The improvement in registration with log-demon is shown in Fig. \ref{fig5}. These results are summarized in Table \ref{tab1} where we show average percentage improvements in 3D volume registration. From these results we observe significant improvements gained by using DDR, especially in reducing the SSD cost, because the optimization has targeted SSD. However, other measures such as SSIM and PSNR have also decreased significantly. Fig. \ref{fig8} shows a residual image (difference image) for the log-demon method with and without using DDR. Significant reduction in residual image intensity is observed when DDR is used. 

\begin{table}[!htb]
\caption{Improvement in registration for IXI dataset: average percentage}
\begin{center}
\begin{tabular}{ |p{0.7cm} | p{0.7cm} | p{0.7cm} | p{0.7cm} | p{0.7cm} |p{0.7cm}|}
\hline
\multicolumn{3}{|c}{DDR with diffeomorphic demons} & \multicolumn{3}{|c|}{DDR with log-demons} \\
\hline
\multicolumn{1}{|c}{SSIM} & \multicolumn{1}{|c}{PSNR} & \multicolumn{1}{|c}{SSD} & \multicolumn{1}{|c}{SSIM}  & \multicolumn{1}{|c}{PSNR} & \multicolumn{1}{|c|}{SSD} \\
\hline

\multicolumn{1}{|c}{9.2} & \multicolumn{1}{|c}{5.0} & \multicolumn{1}{|c}{19.0} & \multicolumn{1}{|c}{11.0}  & \multicolumn{1}{|c}{5.3} & \multicolumn{1}{|c|}{21.0} \\
\hline
\end{tabular}
\end{center}
\label{tab1}
\end{table}
\vspace{-1ex}
\begin{figure}[!htb]
  \centering
	  \hspace{0.6 in}
		
     \includegraphics[width=3.2in, height = 2.3in]{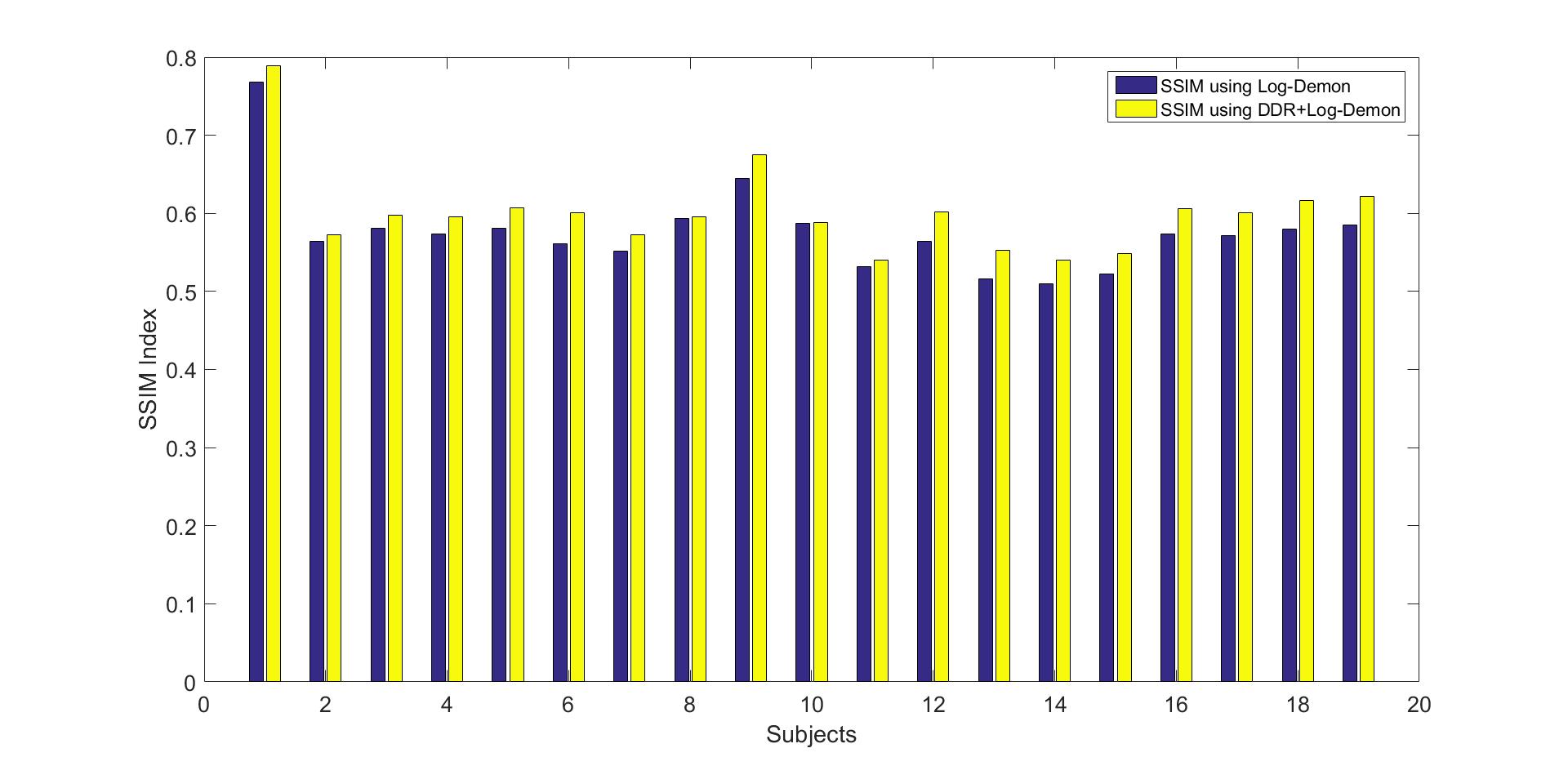}
	
		 \vspace{0.05 in}
		
     \includegraphics[width=3.2in, height =2.3in]{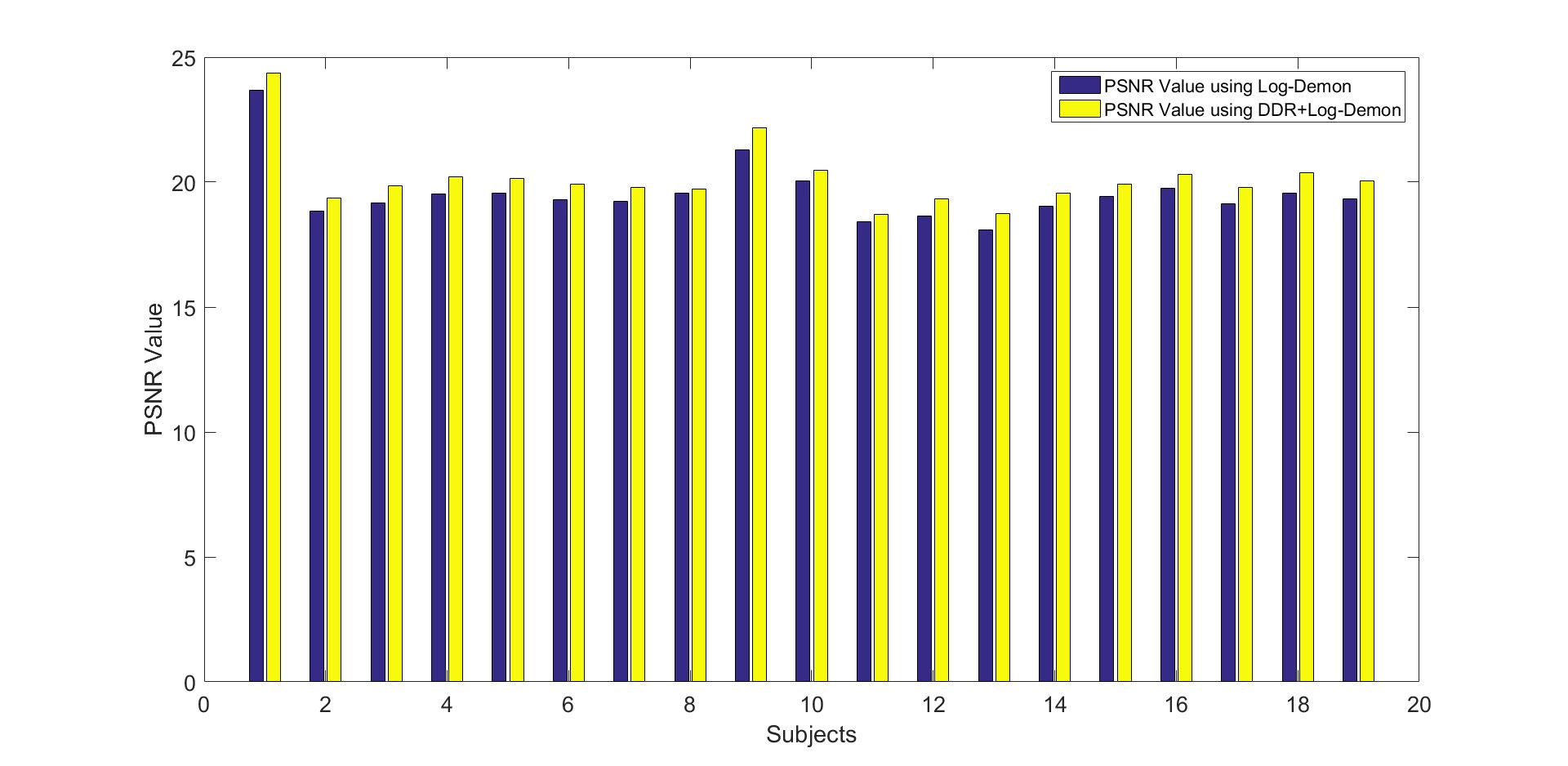}
		 \hspace{0.1 in}
     \includegraphics[width=3.2in, height =2.3in]{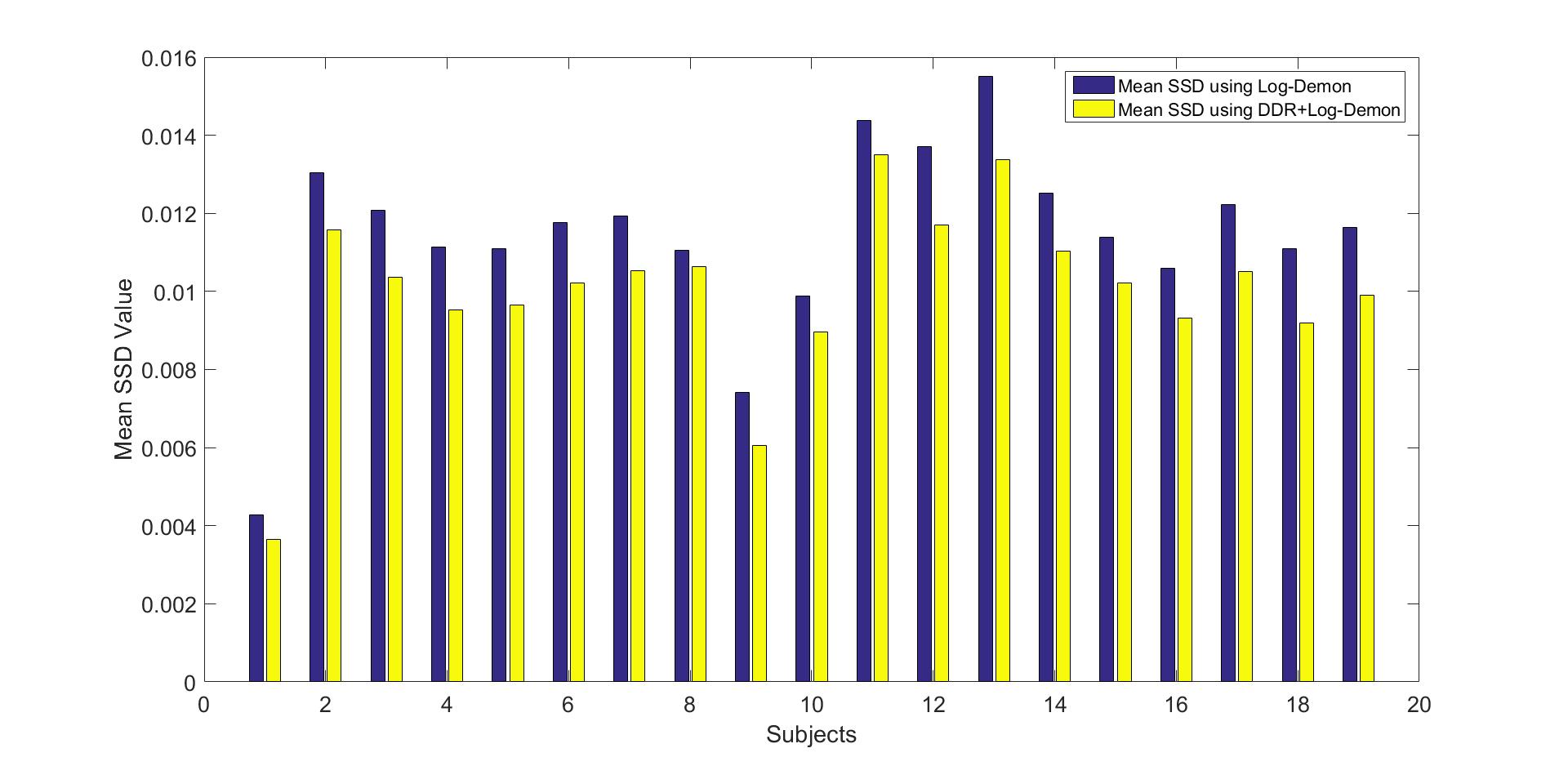} 
     \vspace{0.02in}
\caption{Results on ADNI Dataset. Top : SSIM vs no. of Subjects, Middle: PSNR value vs no. of subjects, Bottom: Mean SSD value vs no. of subjects.}
\label{fig7}
\end{figure}

\subsection{Experiments with ADNI Dataset}
\begin{figure}[!htb]
  \centering
	  \hspace{0.1 in}
		
     \includegraphics[width=3.2in, height = 2.3in]{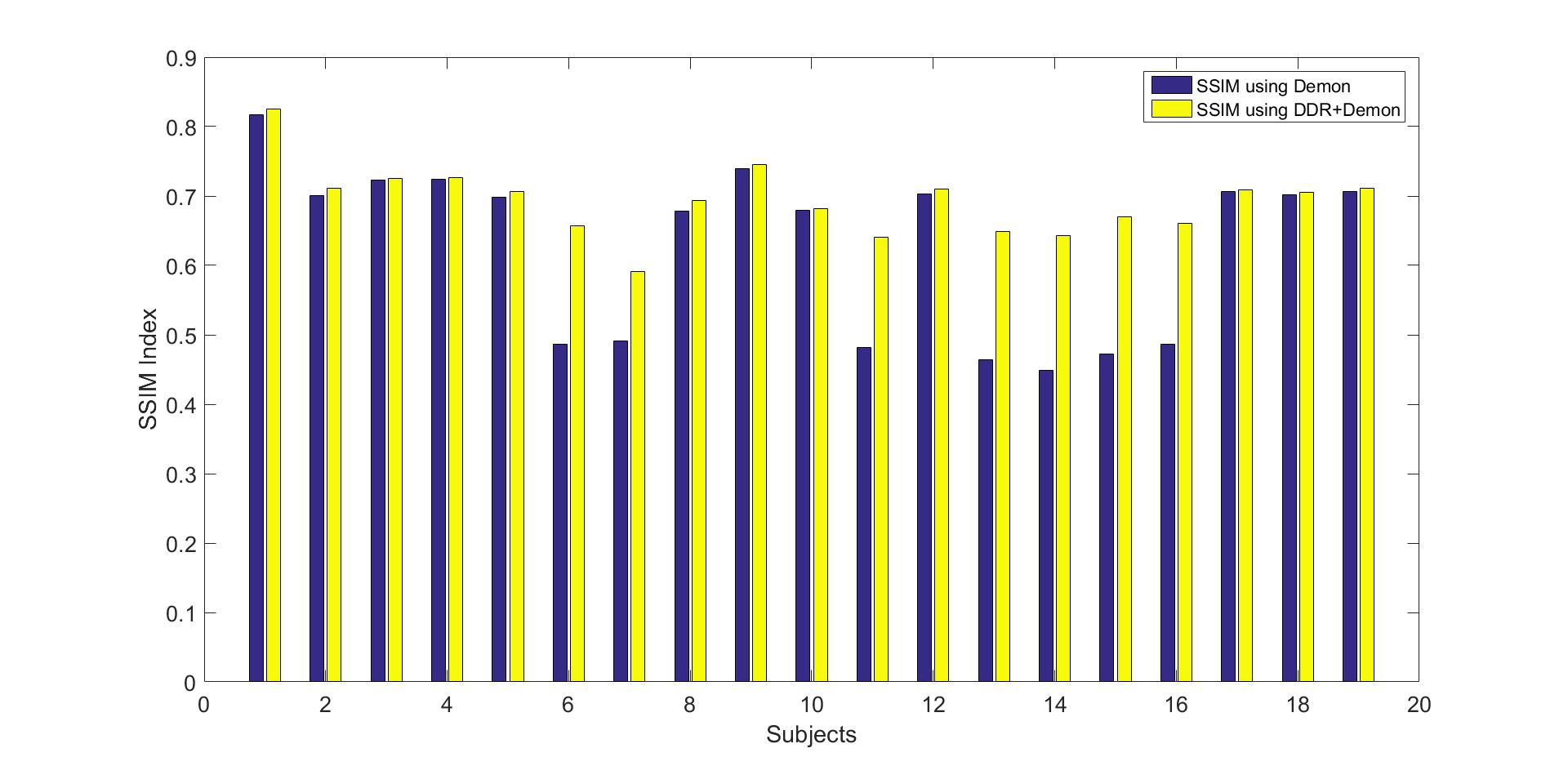}
	
		 \vspace{0.05 in}
		
     \includegraphics[width=3.2in, height =2.3in]{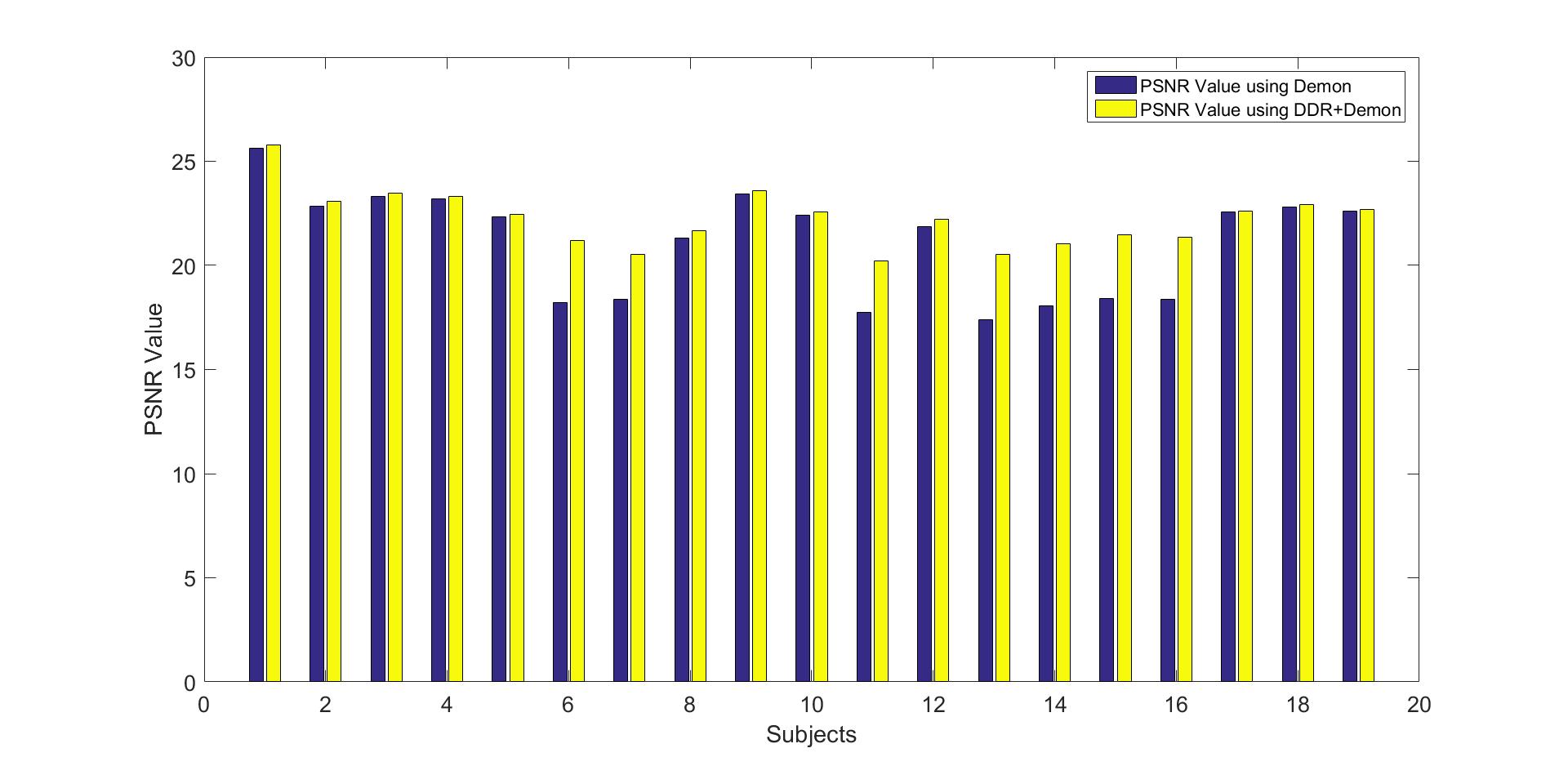}
		 \hspace{0.1 in}
     \includegraphics[width=3.2in, height =2.3in]{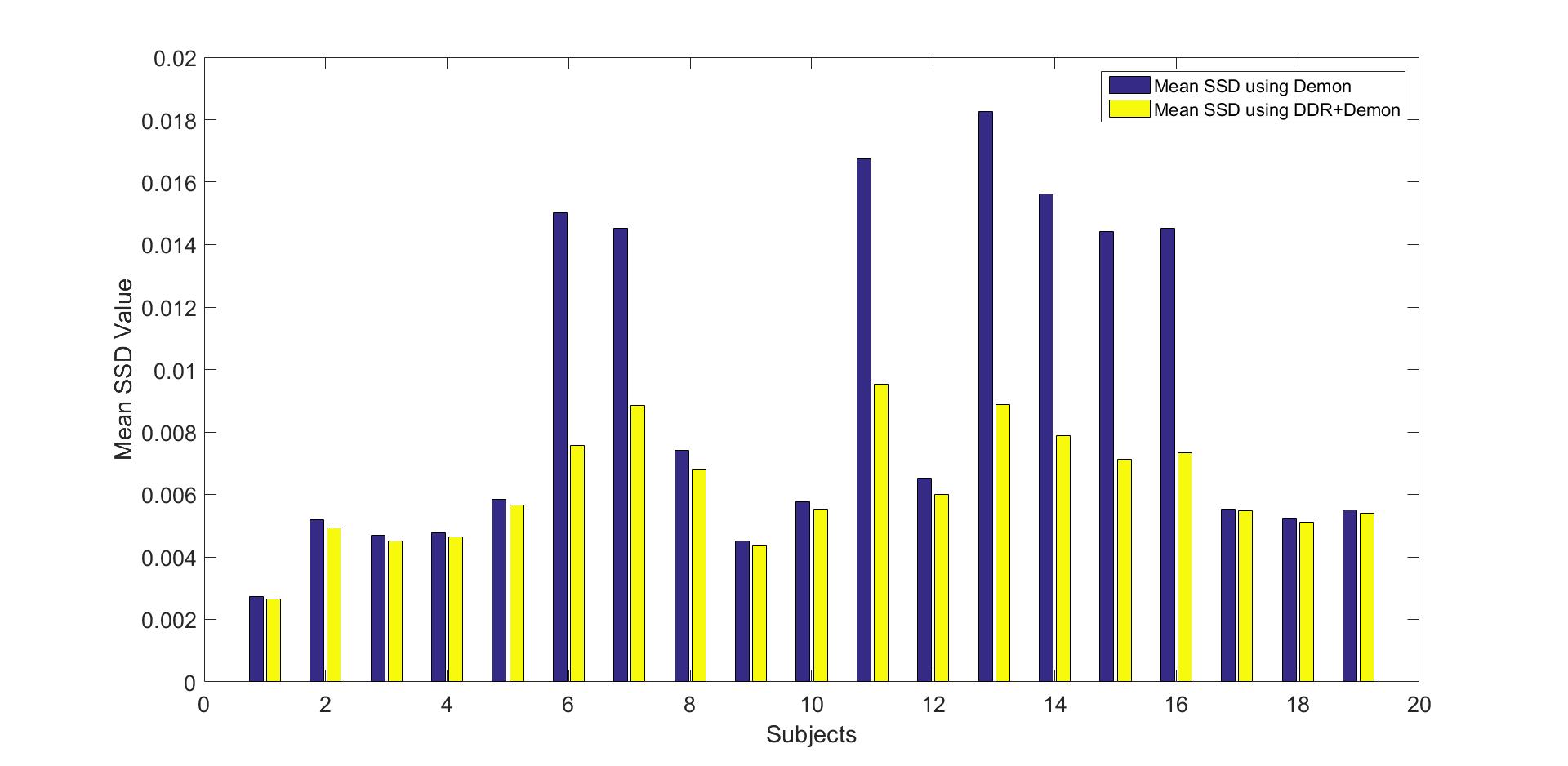} 
     \vspace{0.02in}
\caption{Results on ADNI Dataset. Top : SSIM index vs no. of Subjects, Middle: PSNR value vs no. of subjects, Bottom: Mean SSD value vs no. of subjects.}
\label{fig6}
\end{figure}

\begin{figure}[!htb]
  \centering
     \includegraphics[width=\columnwidth]{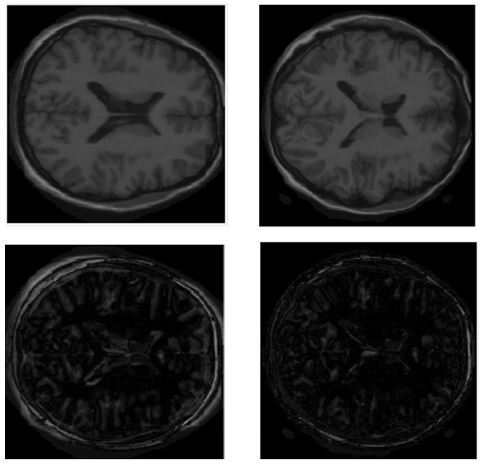}

\caption{Top-left: Registered image using log-demon; top-right: DDR+log-demon result. Bottom-left: Residual image from log-demon. Bottom-right: Residual using DDR+log-demon.}
\label{fig8}
     \vspace{0.1in}
\end{figure}

In these experiments, we have randomly selected 20 MR 3D volumes from the ADNI dataset \url{(http://adni.loni.ucla.edu/)}. Among them, one is randomly selected to be the template, and the rest are registered with it using Diffeomorphic Demons and Log-Domain Diffeomorphic Demons algorithm. The SSIM, PSNR and SSD values are calculated and plotted in Fig.\ref{fig6} and Fig. \ref{fig7}. These improvements are summarized in summarized in Table \ref{tab2}. Once again, we observe significant gains in registration metrics using the proposed DDR.
\begin{table}[!htb]
\caption{Improvement in registration for ADNI dataset: average percentage}
\begin{center}
\begin{tabular}{ |p{0.7cm} | p{0.7cm} | p{0.7cm} | p{0.7cm} | p{0.7cm} |p{0.7cm}|}
\hline
\multicolumn{3}{|c}{DDR with diffeomorphic demons} & \multicolumn{3}{|c|}{DDR with log-demons} \\
\hline
\multicolumn{1}{|c}{SSIM} & \multicolumn{1}{|c}{PSNR} & \multicolumn{1}{|c}{SSD} & \multicolumn{1}{|c}{SSIM}  & \multicolumn{1}{|c}{PSNR} & \multicolumn{1}{|c|}{SSD} \\
\hline

\multicolumn{1}{|c}{13.6} & \multicolumn{1}{|c}{6.2} & \multicolumn{1}{|c}{19.9} & \multicolumn{1}{|c}{4.3}  & \multicolumn{1}{|c}{3} & \multicolumn{1}{|c|}{12.6} \\
\hline
\end{tabular}
\end{center}
\label{tab2}
\end{table}

\section{Conclusions and Future Work}
We have proposed a novel method for improving deformable registration using fully convolutional neural network. While previous studies have focused on learning features, here we have utilized FCNN to help optimize registration algorithm better. On two publicly available datasets, we show that improvements in registration metrics are significant. In the future, we intend to work with other diffeomorphic registration algorithms, such HAMMER \cite{shen07}.

\section*{Acknowledgments}
Authors acknowledge support from MITACS Globallink and Computing Science, University of Alberta.


\bibliographystyle{IEEEbib}
\bibliography{refs}

\end{document}